\documentclass[letterpaper, 10 pt, conference]{ieeeconf}  %

\usepackage{multicol}
\usepackage[utf8]{inputenc} %
\usepackage[T1]{fontenc}    %
\usepackage{hyperref}       %
\usepackage{url}            %
\usepackage{booktabs}       %
\usepackage{amsfonts}       %
\usepackage{nicefrac}       %
\usepackage{microtype}      %
\usepackage{epigraph}
\usepackage{graphicx}
\usepackage{amsmath}
\usepackage{array}
\usepackage{hyperref}
\usepackage[binary-units=true]{siunitx}
\usepackage{subcaption}
\usepackage{bm}
\usepackage{mathtools}
\usepackage{cuted}
\usepackage{etoolbox}
\usepackage{balance}
\usepackage[export]{adjustbox}

\usepackage{wrapfig}

\newcommand\mypara[1]{\vspace{3pt}\noindent\textbf{#1.}}

\usepackage{pgfplots}
\pgfplotsset{compat=newest}
\usetikzlibrary{plotmarks}
\usetikzlibrary{babel}
\usetikzlibrary{arrows.meta}
\usepgfplotslibrary{patchplots}
\usepackage{grffile}
\usepackage{amsmath}
\usepackage{interval}

\usepgfplotslibrary{groupplots}
\usepgfplotslibrary{dateplot}

\IEEEoverridecommandlockouts                              %

\title{\LARGE \bf
Learning Visual Locomotion with Cross-Modal Supervision
}

\author{Antonio Loquercio*, Ashish Kumar*, Jitendra Malik%
\thanks{* denotes equal contribution. All authors are affiliated to UC Berkeley}%
}

\begin{document}

\makeatletter
\g@addto@macro\@maketitle{
  \def\mycolspace{0.95\textwidth}
  \centering
 	\begin{tabular}{c@{\hspace{0.05\textwidth}} c@{\hspace{\mycolspace}}}
 	  \centering
 	  &
\includegraphics[width=0.9\linewidth,trim=0 0 0 0, clip]{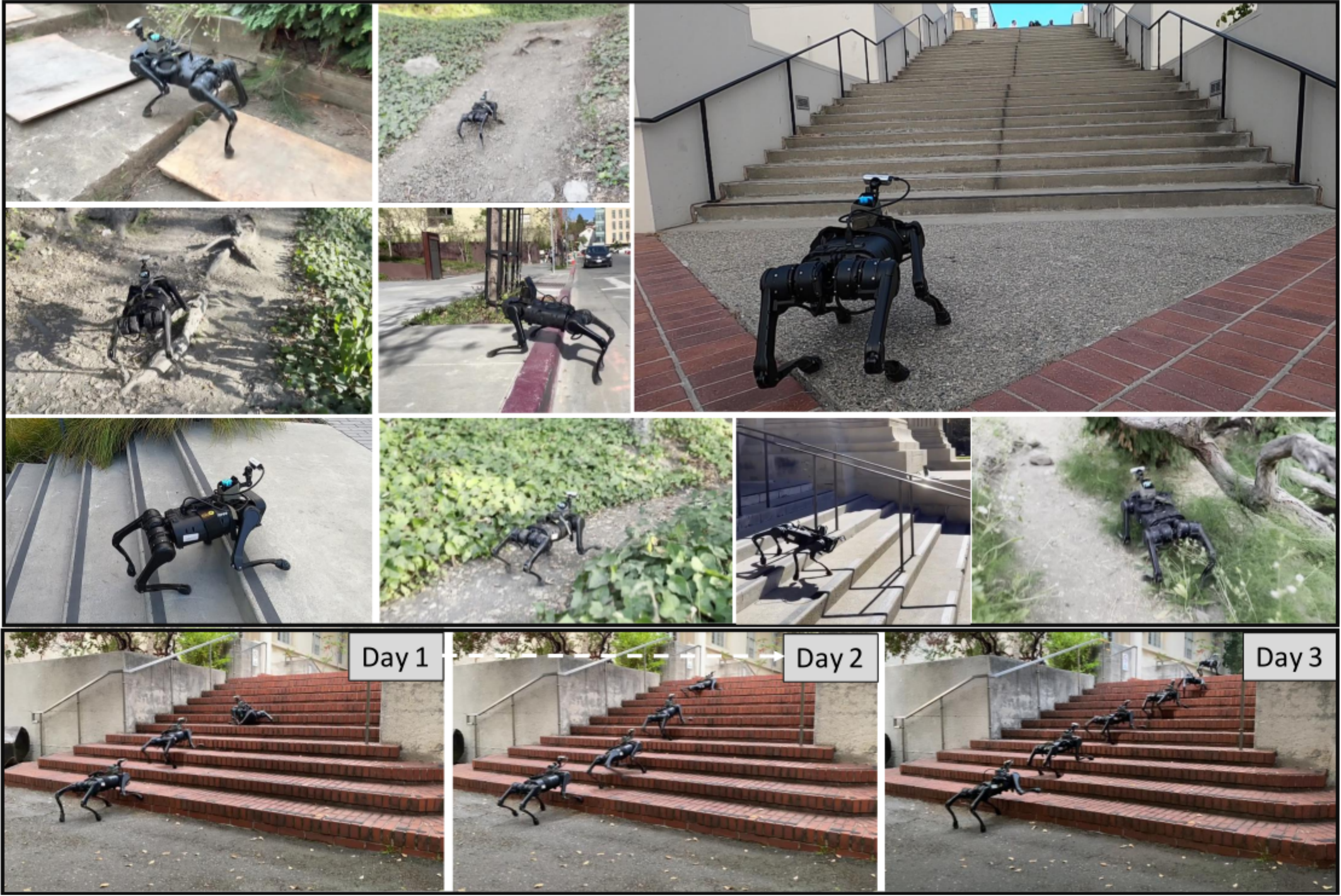} \\
	\end{tabular}
	\captionof{figure}{The deployed walking policy shown above only  uses a monocular egocentric RGB stream and proprioception. The terrains include stairs (up to 19cm high), curbs (up  to 20cm high), slopes (35$^\circ$), dirt roads, and unstructured construction sites. Several of these require precise foothold placement which is achieved by predicting the upcoming terrain using a visual lookahead module. This module is trained entirely in the real world. To do so, our proposed algorithm Cross-Modal Supervision(CMS) uses onboard proprioception to supervise a vision module. This naturally allows the policy to continually learn in the real world with its own experience. We show one such progression of continual learning at the bottom, where the policy goes from an initial success rate of 40\% to 100\% with less than 30 minutes of real-world data. Video results and code at \url{https://antonilo.github.io/vision_locomotion/}.}
	\vspace{-0.5cm}
	\label{fig:fig1}
}
\makeatother
\maketitle
\maketitle
\thispagestyle{empty}
\pagestyle{empty}

\begin{abstract}

In this work, we show how to learn a visual walking policy that only uses a monocular RGB camera and proprioception. Since simulating RGB is hard, we necessarily have to learn vision in the real world. We start with a blind walking policy trained in simulation. This policy can traverse some terrains in the real world but often struggles since it lacks knowledge of the upcoming geometry. This can be resolved with the use of vision. We train a visual module in the real world to predict the upcoming terrain with our proposed algorithm Cross-Modal Supervision (CMS). CMS uses time-shifted proprioception to supervise vision and allows the policy to continually improve with more real-world experience. 
We evaluate our vision-based walking policy over a diverse set of terrains including stairs (up to 19cm high), slippery slopes (inclination of 35$^{\circ}$), curbs and tall steps (up to 20cm), and complex discrete terrains. We achieve this performance with less than 30 minutes of real-world data.
Finally, we show that our policy can adapt to shifts in the visual field with a limited amount of real-world experience.
\end{abstract}

\setcounter{figure}{1}  

\section{INTRODUCTION}
\begin{figure*}[t]
    \centering
    \includegraphics[width=0.95\linewidth]{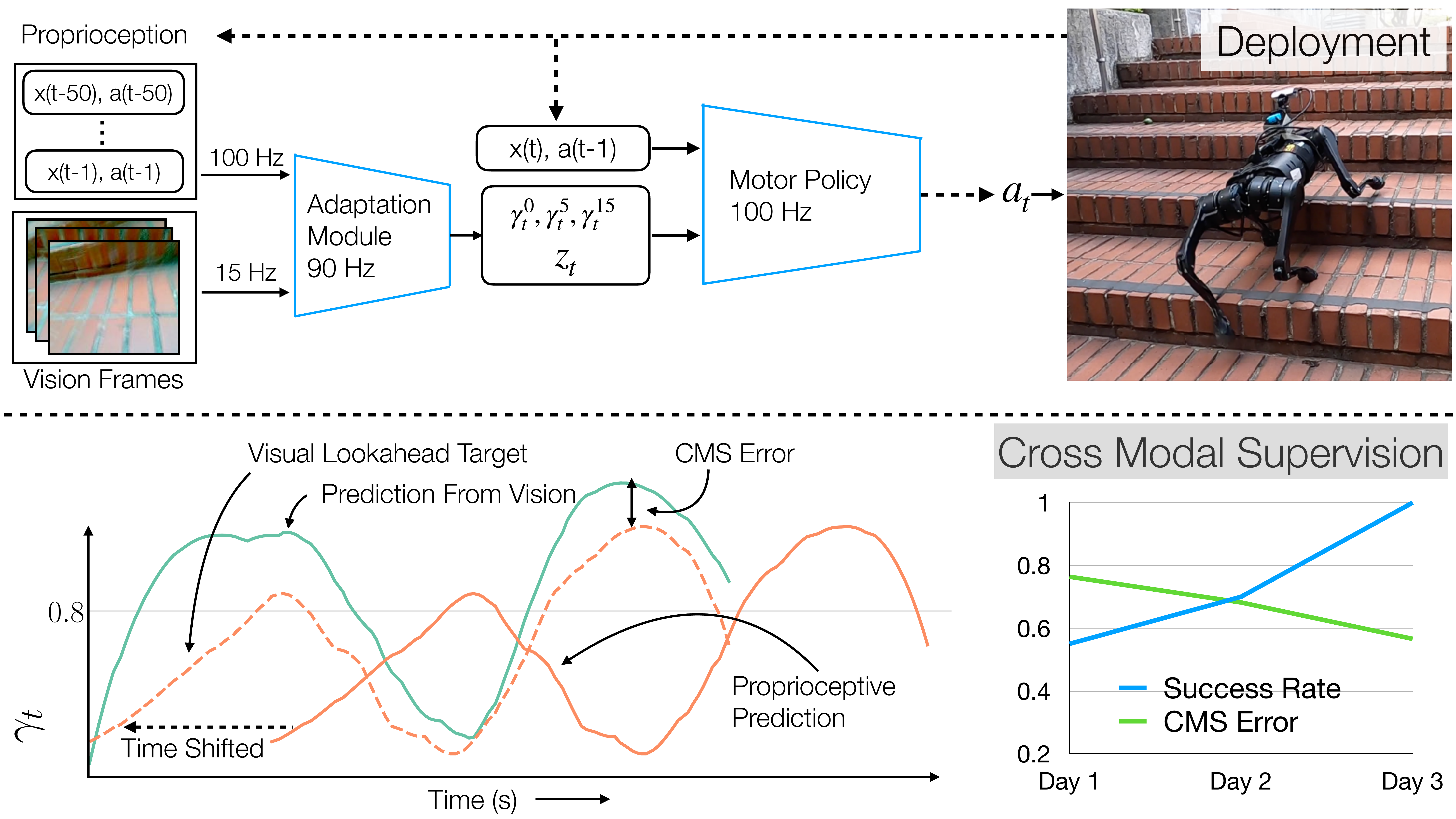}
    \caption{\textbf{Top:} During deployment, we use an adaptation module to predict the extrinsic parameters $z_t$ and an estimate of the terrain geometry below, 5cm, and 15cm in front of the robot ($\gamma^{0}$, $\gamma^{5}$, $\gamma^{15}$) from egocentric RGB and proprioception. \textbf{Bottom:} We train the visual module in the real world to predict a time-shifted proprioceptive estimate of the terrain's slope by minimizing the CMS Error. We observe that CMS enables the adaptation module to improve continually with more real-world experience. On the bottom right we see that as the CMS error decreases the walking policy improves.}
    \label{fig:method} 
    \vspace{-0.5cm}
\end{figure*}

Gibson famously said that ``We see in order to move and we move in order to see". For land animals, if we interpret moving as walking, we might say ``we see in order to walk and we walk in order to see". Taken literally, this statement is surely false because obviously blind humans can walk. Blind robots can walk too~\cite{kumar2021rma,lee2020learning}. However Gibson wasn't completely wrong either -- while blind people can walk, they find it challenging to walk, and cannot walk as fast on complex terrains such as stairs. They have to probe gently with a walking stick first before they can start walking safely. Research on human vision ~\cite{patla1997understanding, patla2003strategies,loomis1992visual, bonnen2021binocular} shows that vision gives us "look-ahead" - when we can see the ground ahead of us, our foot placement is smoother and the walking more efficient. In addition, such look-ahead is tightly coupled to the walking strategy~\cite{Kretch_2013}. We study this question in a robotics context. This paper is about our development of walking policies for quadruped robots, using the visual system for look-ahead-- predicting the height of the terrain in front of the robot before it gets there. Unsurprisingly, we found that policies with look-ahead indeed do better than blind ones (see Figure~\ref{fig:fig1}), thus validating the spirit, if not the letter of Gibson's statement.

A vision sensor in a robotics context could be a single RGB camera, or one of the many different devices directly capturing depth using stereopsis or LIDAR for instance.  Our decision was to go for RGB cameras since (1) there is evidence from human vision that stereo input contributes relatively little to making walking faster/more efficient~\cite{bonnen2021binocular} (2) for objects in terrestrial perspective, priors enable one to make reasonable depth estimates even from a single image~\cite{Ranftl2022} (3) the multiple views captured by a moving robot could in theory be exploited to recover depth directly as well~\cite{hartley2003multiple}. We don't wish to argue our point too strongly -- there has been  passionate but not necessarily productive debate on this in the context of self-driving cars already -- but  just say that the RGB version is scientifically interesting and there might also be engineering contexts where it is favored (cost, power, passive sensing). From a research novelty perspective, there are several systems which exploit depth input for guiding walking~\cite{agarwal2022ego,miki2022learning}, whereas the RGB only design choice has hardly been studied.

The RGB only choice leads to a technical challenge. The most popular techniques for training walking policies with RL  typically train in simulation and then deploy zero shot in the real world~\cite{kumar2021rma,lee2020learning}. This works out fine with depth inputs because the sim-to-real gap for depth is small, and multiple systems have been successfully demonstrated for walking and flying~\cite{loquercio2021learning,agarwal2022ego,miki2022learning}. However rendered RGB images in a simulator look very different from those in the real world. The higher the fidelity  desired in the rendering, the greater the computational cost, which makes it practically infeasible to use in an inner loop for training  RL policies because of their high sample complexity. 

Our core insight is that we can train the vision part of our policy in the real world from onboard sensors, while still managing to train the action policy in a simulator. In a simulator we can train policies which don't use vision but "cheat" - they have access to look-ahead information of terrain height at a few "look-ahead" points ahead of the robot - and then it is a separate problem to train a vision system from real world experience which can predict the terrain height at those look-ahead points. This vision system is trained using cross-modal supervision (CMS) from proprioception with a time lag - I see a point A ahead of me whose height is currently unknown, but when my feet get to A, then its height can be inferred from the joint angles of my body. There is a "chicken-and-egg" aspect to this - to develop a vision system which can provide look-ahead for walking, you need to be able to walk first to get the data to provide the supervision for the vision system! Fortunately, there is no infinite regress problem here, as the robot can walk (clumsily of course) when blind, so the bootstrapping can start from there.

Experiments show that our policy can walk over a diverse set of terrains including stairs (up to 19cm high), slippery slopes (inclination of 35$^{\circ}$), curbs and tall steps (up to 20cm), and complex discrete terrains from a single onboard RGB camera and proprioception. To systematically understand the improvements over the initially deployed policy, we take 4 real-world stair cases on which the initial blind policy starts off at around 50\%, and compare the performance of the visual walking policy as it improves with experience. We find that in all these cases, the visual walking policy reaches 100\% performance with less than 30 minutes of data, collected across 4 days.(Figure~\ref{fig:rw_improvement} and Section~\ref{sec:results}).

Since the proposed algorithm, CMS (Cross-Modal Supervision), enables learning in the real world from onboard sensors our system can continue to learn lifelong. The experience collected online is used continually to improve the visual system, which in turn, improves the performance of the overall policy.
We systematically evaluate this aspect of CMS in 4 challenging setups in the real world and show significant and consistent improvements in performance with less than 7 minutes of data each day. 
We also analyze the qualitative change in behavior of the visual terrain predictor compared to the proprioceptive estimator (Section~\ref{sec:results}) and see precise foot placement with vision.
Finally, we perform the widely-known prism experiment~\cite{barrett2012prism} to demonstrate the visual plasticity of the learned model: our experiments indicate that with very limited data the model can adapt to large shifts in the visual field.
Video results can be seen at \url{https://antonilo.github.io/vision_locomotion/}.

\section{Overview}

We want a visual walking policy that can walk on complex terrains. We train adaptive walking policies similarly to RMA~\cite{kumar2021rma}, but with the following significant change -- we have an additional module that estimates the upcoming terrain geometry from the last three egocentric frames. More concretely, the policy $\pi$ takes a) an estimate of terrain in front and below the robot $\bm{\gamma}$ for visual adaptation to complex terrain, b) an estimate of the environment parameters $z$ to enable adaptation to variations in the environment (similar to RMA~\cite{kumar2021rma})) the current proprioceptive state $\bm{x}$, and uses them to predict the target actions $\bm{a}$. This system is shown in the top part of Figure~\ref{fig:method}. Once we have such a system deployed, it collects egocentric visual data, as well as proprioception. This multi-modal data can be used for Cross-Modal Supervision (CMS) to continually improve the visual estimates of the terrain parameters $\bm{\gamma}$. This is enabled by the key insight that proprioception can get an accurate estimate of the same terrain that the egocentric camera records ahead of time.
This allows using proprioceptive estimates of terrain to supervise the visual estimate, as shown in the bottom half of Figure~\ref{fig:method}. 
Specifically, we train a vision module to predict the future proprioceptive estimate of terrain from the egocentric visual input (that looks in front of the robot) by minimizing the CMS error. Since the supervision comes from onboard sensors, we continually do so to get an increasingly better policy with more experience. 

We initialize the CMS improvement procedure from a blind policy $\pi_{blind}$, which can be trained in simulation and transferred to the real world. This policy has no visual lookahead and uses proprioception to estimate the terrain under the feet of the robot as the robot walks. Once we have this policy, we can deploy it in the real world, and train a module for visual terrain lookahead using CMS. Once we have this visual predictor of the upcoming terrain, we can then use the lookahead policy $\pi$ (also trained in simulation) to get a visual walking policy in the real world.

\begin{figure*}
    \centering
    \includegraphics[width=0.95\textwidth]{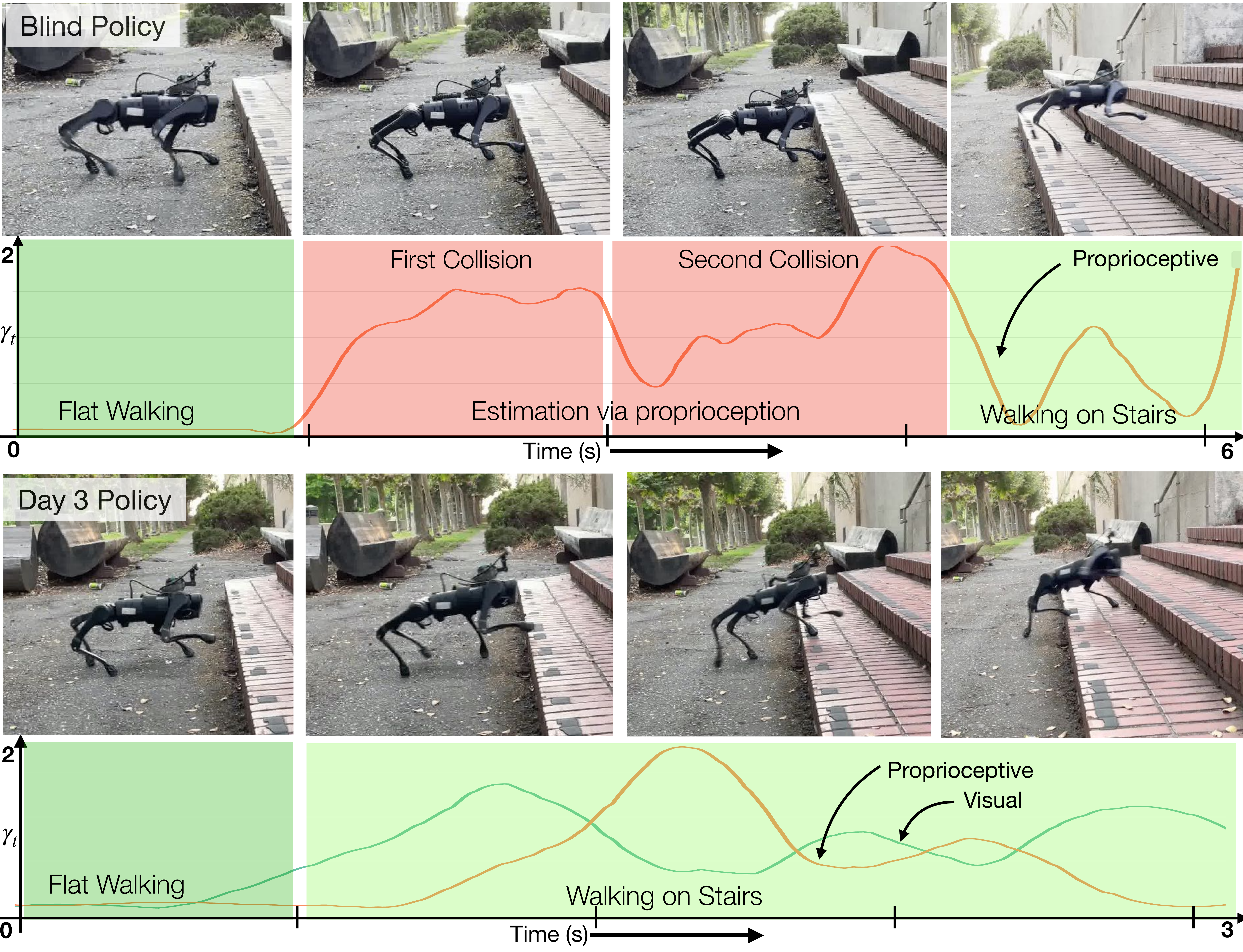}
    \caption{\textbf{Behavior Analysis}: The top row shows a blind policy, which needs to touch the stairs multiple times before correctly estimating its height and climbing over it. In contrast, a vision-based policy trained to predict the future proprioceptive estimate of terrain can anticipates changes in the terrain geometry and adapt its gait accordingly. This speeds up the initial climb by approximately two times and gives an overall higher success rate.}
    \label{fig:analysis}
    \vspace{-0.5cm}
\end{figure*}

\begin{figure*}
    \centering
    \includegraphics[width=0.95\textwidth]{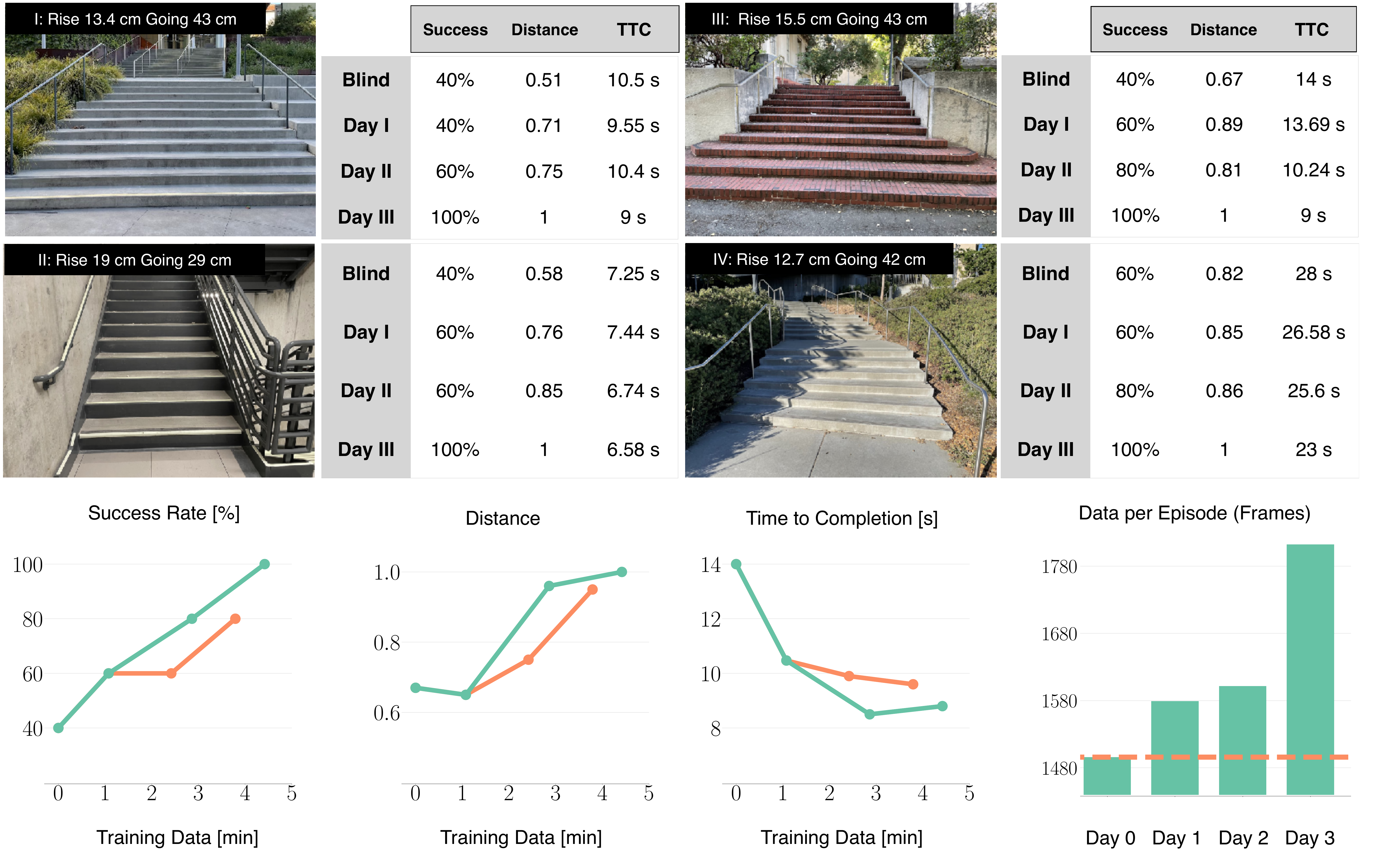}
    \caption{\textbf{Real-World Improvement}: (Top Row) We find that using CMS continually improves the performance of the walking policy, from approximately 50\% to 100\% in all the settings considered above. This is enabled by the use of vision, which is trained in the real world with CMS. The final policy uses less than 30 minutes of data collected over 4 days in different conditions. (Bottom Row) We observe that using the latest available vision policy (green) to collect data, instead of the initial blind policy (orange), leads to faster convergence of the final visual policy. This shows that the data collected with the latest visual policy is of higher quality. Moreover, we observe that since the vision policy can walk longer without falling, it significantly improves the efficiency of data collection in the real world.}
    \label{fig:rw_improvement}
    \vspace{-0.5cm}
\end{figure*}

\section{Method}

\subsection{Simulation Training}
\label{sec:base_policy}

We train two walking policies ($\pi$ and $\pi_{blind}$) in simulation to walk on complex terrains. Both the policies take the state $\bm{x}_t$, the extrinsics vector $\bm{z}_t$, and the terrain information $\bm{\gamma}_t$ to predict the target joint angles $\bm{a}_t$ at 100Hz which is converted to torques using a PD controller. The lookahead policy $\pi$ takes an additional input $\gamma_{t+\Delta t}$, which is the terrain estimate at 15cm lookahead in front of the robot. Concretely, 

\mypara{State Input} $\bm{x}_t = [\bm{q}_t, \bm{q}'_t, \bm{a}_{t-1}, \bm{z}_t, \bm{\gamma}_t]^\top$, where  $\bm{q}_t \in \mathbb{R}^{12}$ is the current joint position, $\bm{q}'_t \in \mathbb{R}^{12}$ the joint velocity, $\bm{a}_{t-1} \in \mathbb{R}^{12}$ the previous action.

\mypara{Extrinsics Vector} $\bm{z}_t \in \mathbb{R}^{8}$ is a latent representation of the environment parameters which includes payload, motor strength, binarized foot contacts, linear velocity and friction.

\mypara{Terrain Geometry Vector} $\bm{\gamma}_t$ is a latent representation of the ground truth terrain geometry $\bm{h}_t$ under the feet of the robot, and $\bm{\gamma}_{t+\Delta t}$ is the terrain estimate 15cm infront of the robot. 
This gives us:
\begin{align}
    \bm{z}_t &= \mu(\bm{e}_t) \\
    \bm{\gamma}_t &= \delta(\bm{h}_t)\\
    \bm{\gamma}_{t + \Delta t} &= \delta(\bm{h}_{t + \Delta t})\\
    \bm{a}_t &= \pi(\bm{x}_t, \bm{z}_t, \bm{\gamma}_t, \bm{\gamma}_{t + \Delta t})\\
    \bm{a}_t &= \pi_{blind}(\bm{x}_t, \bm{z}_t, \bm{\gamma}_t),
\end{align}
where $\mu$, $\pi$ and $\pi_{blind}$ are MLPs with two hidden layers and $[256, 128]$ and $[128,128]$ neurons, respectively. Note that we use the same network $\delta$ (hidden dimensions of $[64,16]$) to process $\gamma_t$ and $\gamma_{t+\Delta t}$. 
We train the $\pi_{blind}$ and the dependent encoders jointly in an end-to-end manner using model-free reinforcement learning. Once trained, we freeze the network $\delta$ and $\mu$, and train policy $\pi$ using model-free RL. We now explain the environment design and the reward structure we use for RL which is shared between both the policies.

\mypara{Environment Design} We recreate in simulation a set of natural conditions to elicit robust locomotion. Specifically, we train on fractal terrains (similar to ~\cite{kumar2021rma,fu2021minimizing}) and parameterized inclines and stairs, which typically represent a majority of the commonly encountered terrains in human-made environments. 
To design the distribution of staircases, we take inspiration from a classic reference in the field of architecture~\cite{staircase_design}.
According to the latter, to be comfortable for humans, staircases should have height in the range $[10,19]$cm and lengths not smaller than $30$cm.
In addition, two heights and one length should make for the length of a step, which is approximately $1$mt.
We use these two recommendations to build our parameterized staircases: we uniformly sample step heights in the range $[10,21]$cm and lengths from the set $[30,40,50,60]$cm.

\mypara{RL Rewards} We design a reward function to promote the agent to move with a user-defined forward and angular speed in the ranges ${v_x}^d \in [0,0.5]$m/s and ${w_z}^d \in [-0.4,0.4]$ rad/s, while penalizing lateral speed and jerky motions.
We define $\bm{v}$ as linear velocity, $\bm{\omega}$ as the angular velocity, $\bm{\alpha}$ and $\dot{\bm{\alpha}}$ as joint angles and velocities, $\bm{\tau}$ as joint torques, $\bm{v}^f$ the feet velocity and $g$ the binary foot contact indicator.
Accordingly, we define the reward at time $t$ as the sum of the following terms:
\begin{itemize}
    \item Forward: $min({v_x}^d,v_x)$
    \item Lateral: $\|{v_y}\|$
    \item Angular: $-\|{w_z}^d - w_z\| + {w_x}^d$
    \item Work: $-\|\bm{\tau}^\top \cdot (\bm{\alpha}_t - \bm{\alpha}_{t-1}) \|$
    \item Foot Slip: $-\|\text{diag}(g)_t \cdot \bm{v}^f_t \|$
\end{itemize}
The scaling of each of the previous factors is $65$, $1$, $40$, $0.05$, $0.2$. We additionally add a survival bonus of $9$, which is doubled whenever the robot is tasked with walking on a staircase.

\mypara{Curriculum} 
We start by training exclusively on flat terrain for $500$M steps. Afterwards, we train on progressively difficult staircases: starting from heights of $10$cm and gradually increasing the step height every $100$M steps until it reaches the maximum. 
During training on staircases, we sample flat terrains with $10\%$ probability to avoid catastrophic forgetting.
To guide the optimization, we additionally add a regression loss to match the actions from a policy exclusively trained on flat(~\cite{fu2021minimizing})
We randomly sample the environment parameters such as payload, friction, and motor strength from the ranges reported in~\cite{kumar2021rma}.

\mypara{Sim to Real Transfer} To transfer policy $\pi_{blind}$ to the real world, we train a network $g_0$ to predict $\bm{z}_t$ and $\bm{\gamma}_t$ from proprioception and action history.
We use the same structure and training procedure for $g_0$ as proposed by RMA~\cite{kumar2021rma}.

\subsection{Vision-Based Locomotion via CMS}

Once $\pi_{blind}$ is deployed in the real world, we collect a dataset of trajectory rollouts in the real world $\mathcal{D}={([I_t,\bm{x}_t], \gamma_{t})}$, where $I_t$ is the egocentric visual input from the onboard camera, $\bm{x}_t$ is the current proprioceptive state and $\gamma_{t}$ is the current geometry estimate. We can now use this dataset to train a convolutional neural network $g_i$ to estimate the current extrinsics $\bm{z}_t$ and the future terrain geometry $\bm{\gamma}_{t+\Delta t}$ using supervised learning as shown in Figure~\ref{fig:method}. Having a visual predictor of upcoming terrain we can now use the lookahead policy $\pi$ which has better performance than the blind policy. 
\subsection{Lifelong Learning via CMS}

CMS naturally allows for lifelong learning.
Since we continuously collect the dataset $\mathcal{D}$ with real-world experience, we can use the $\gamma$ estimate from proprioception to continuously train the visual lookahead terrain predictor during execution and improve performance.

\section{Experimental Setup}
\label{sec:results}

\textbf{Simulation Setup}: We use the RaiSim simulator~\cite{raisim} for
rigid-body and contact dynamics simulation. We import the
A1 URDF file from Unitree.
Each RL episode lasts for a maximum of 1200 steps with early termination if the roll or pitch exceeds a threshold, or the base of the robot is very close to the ground. 
The control frequency of the policy is 100 Hz, and the simulation time step is 0.025s.

\mypara{Vision-Based Estimator} We build a custom model to predict the future $\gamma$ to allow inference on limited compute.
The input of the model are the latest three frames (camera frequency is 15Hz), converted to grayscale and concatenated to make an input tensor.
We compute features from such tensor with a shufflenet-V2 model~\cite{ma2018shufflenet}.
We use as features the last layer before global-average pooling to maintain spatial information.
The features are then projected to a 2dim channel space with 1D convolutions and processed by an MLP with hidden dimension of $[128,64]$ to predict $\gamma$.
We additionally provide a history of 50 IMU measurements (roll and pitch) and desired velocity commands to the predictor.
This network outputs a $128$ dimensional embedding which is later concatenated with the shufflenet features and passed through the final MLP.

\section{Results and Analysis}

\begin{figure*}
    \centering
    \includegraphics[width=0.95\textwidth]{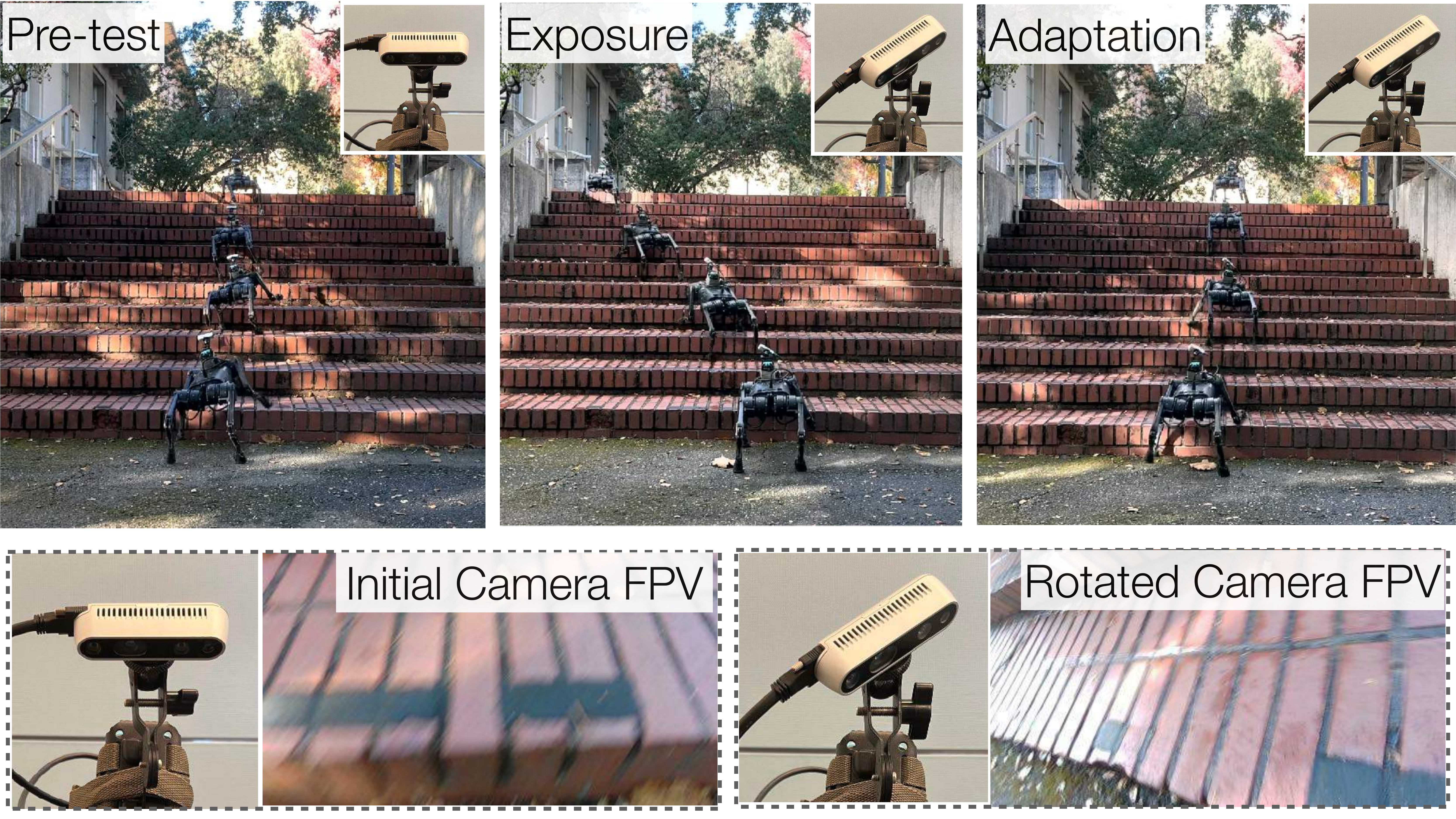}
    \caption{\textbf{Prism-Adaptation Test}: We perform the widely-known prism test~\cite{von1925helmholtz} on our robot. In this test, the visual field of the subject is altered by using a prism in front of the eyes. To simulate this, we rotate the camera from its nominal position.
    (Bottom Row) The left camera is the nominal position, and the right camera is the one after rotation. The view from the camera is shown next to the camera orientation. Despite the robot's position being the same, the visual appearance changes drastically between the two camera configurations.
    (Top Row) The left image shows the visual walking policy under nominal conditions (\emph{pre-test}), the center image is the performance of this visual walking policy immediately after the camera is rotated (\emph{exposure}), and the right image shows the performance after training for 80 seconds of experience (\emph{adaptation}). The behavior recovers with just 1 minute of finetuning on the new visual input.}
    \label{fig:prism_adaptation}
    \vspace{-0.5cm}
\end{figure*}

We design an evaluation procedure to answer three main questions:  (i) What are the quantitative and qualitative benefits of vision-based locomotion over blind locomotion in the real world? (ii) During the continual training in the real world, what benefit does deploying the latest available policy have over collecting all the data with a blind policy and using it at once? (iii)  What is the quantitative benefit of having a visual lookahead on terrain geometry?

\subsection{Lifelong Vision Learning}

A blind policy walks very differently than a vision-based policy.
In Figure~\ref{fig:analysis} we analyze the qualitative change in behavior as a consequence of improved terrain prediction by plotting the estimated $\gamma$ as a function of time for the blind policy and the visual walking policy.
The blind policy is clumsy but "exploratory", tapping the terrain twice before walking on it.
Conversely, the vision policy can anticipate the geometry of the terrain and adapt its stride accordingly.
This results in a smooth and agile gait, enabling the robot to start climbing the stairs without stumbling and speeding up the initial climb by approximately two times.

The quality of the vision predictor improves with experience.
We study the dependence between experience and performance in Fig.~\ref{fig:rw_improvement}.
We select four stairs of different heights and lengths and evaluate the performance over multiple iterations of data collection.
For each policy, we perform five trials on each stair. We define a "day" as the policy trained with all data collected by its predecessors.
All the metrics constantly improve as a function of time.
A policy trained on three "days" of experience, or more specifically on 26 minutes of walking data, can climb all stairs without a single failure.
Overall, the vision policy can walk faster (average speed is 26\% higher) and more effectively (the success rate is 50 percentage points higher) than its blind counterpart.

Since the visual walking policy can go longer without falling it collects data more efficiently. 
As shown in Fig.~\ref{fig:rw_improvement}, the amount of experience collected on average during a real-world experiment increases by 21\%.
In addition, the quality of the data collected using vision is of superior quality.
For approximately the same amount of data, the policy trained on data collected exclusively from the blind controller performs worse than the one trained in an iterative fashion.
Intuitively, this is due to the fact that the blind controller occasionally stumbles or taps the terrain before taking a step, which results in corrupted data.
Conversely, a (possibly intermediate) vision controller is more precise in its gait, leading to higher quality data.

\mypara{Generalization Experiments} We test the generalization of our final vision policy on several challenging terrains, as shown in Figure~\ref{fig:fig1}.
The terrains include previously unseen staircases, steep inclines (up to 35 degree steep), discrete terrains, and curbs.
Our approach was successful on 80\% of previously unseen staircases, 100\% of curbs, and 60\% percent of inclines, which is remarkable since the vision module was never trained for them.
All generalization videos are available at \url{https://antonilo.github.io/vision_locomotion/}.

\mypara{Visual Plasticity} The prism test is a widely-known experiment designed to test adaptation to an artificial shift of the visual field~\cite{von1925helmholtz}.
This test is generally performed on patients to improve the spatial deficits caused by brain damage, e.g., after a stroke~\cite{barrett2012prism}.
A prism adaptation session consists of three phases: the \emph{pre-test}, where a subject performs a task without any disturbance of the visual field; the \emph{exposure}, where the subject performs the same task under an horizontal shift of the visual field (generally wearing prism wedges); and the \emph{adaptation}, where the subject adapts to the new visual field and can perform the task at pre-test levels.
Ideally, the subject should complete the task perfectly in the pre-test phase and quickly adapt to the misalignment of the visual and proprioceptive maps caused by the prism wedges.

We perform a similar test to measure the adaptation ability of our visual locomotion policy.
However, instead of adding prism wedges, we shift the visual field by rotating the camera on its yaw axis of approximately 30 degrees (Fig.~\ref{fig:prism_adaptation}).
Such shift makes the terrain in front of the robot almost unobservable. %
Immediately after altering the robot's visual field, the policy stumbles on the steps and accumulates drift while climbing stairs.
To account for such vision impairment, we fine-tune the vision policy by minimizing the CMS error.
Specifically, after each trial in the modified setup, we finetune only the last 3 layers of gamma predictor for 10 epochs.
After just three trials (roughly corresponding to 80 seconds of data), the vision policy learns to account for the  systematic bias in the visual field and manages to correctly anticipate stairs and walk straight (Fig.~\ref{fig:prism_adaptation}).

Finally, we bring the robot's camera back to its initial position and repeat the task.
In this phase, generally referred to as \emph{post-test}~\cite{barrett2012prism}, the robot is required to quickly adjust back to its original visual field.
We repeat the same procedure as above and observe that also in this case, after two trials, the policy adapts back to its original visual field.
We invite the reader to visit our project web-page for a video of the experiment.

\subsection{The value of visual lookahead}

We compare in simulation the performance of policies with access to a larger ground-truth terrain lookahead. Table~I shows the results of this evaluation.
These results indicate a sharp increase in performance (from 58\% to 75\%) when look ahead increases from 0 to 15cm.
\begin{table}[h]
\centering
\scalebox{0.8}{
\begin{tabular}{ccccc}
\toprule
Look Ahead & Success ($\uparrow$) & TTF ($\uparrow$)  & Distance (m) ($\uparrow$)  & Smoothness ($\downarrow$) \\
\midrule
0cm (RMA~\cite{kumar2021rma})  & 58\%    & 0.83  & 4.11         & 34.19              \\
5cm  & 64\%    & 0.88   & 4.67          & 44.30            \\
15cm & 75\%    & 0.95   & 4.73         & 30.77                \\
25cm & 74\%    & 0.96   & 4.96          & 37.19               \\
35cm & 74\%    & 0.94& 4.66          & 31.43              \\
\bottomrule
\end{tabular}
}
\caption{Performance of motor policies with access to different ground-truth visual lookahead. The value of lookahead saturates after 15cm.}
\vspace{-0.5cm}
\end{table}

\section{Related Work}

\subsection{Blind Locomotion}
Traditionally, locomotion controllers were model-based~\cite{buchli2009compliant, Ames_2014, Khoramshahi_2013, Hyun_2016, bellicoso2017dynamic, kalakrishnan2011learning, Byl2009_dynamically}.
However, these methods require accurate modeling of the robot and substantial task-specific engineering for tuning locomotion gaits and behaviors.
While some of these problems could be mitigated with optimization-based controllers~\cite{gehring_2016ram, Calandra_2015}, these methods tend to become brittle outside of lab-controlled conditions~\cite{hwangbo2019learning,lee2020learning}.
While using learning for locomotion has a long history~\cite{Zico_Kolter_2011, learning_zucker_2011}, recent deep reinforcement learning (RL) controllers achieved impressive results in real-world rough terrains~\cite{lee2020learning, kumar2021rma}.
Bbeing proprioception relatively low dimensional, training blind controllers can be done within minutes in highly optimized simulators~\cite{rudin2022learning}.
Independently of the training procedure, blind controllers can only react to varying terrain conditions, but not anticipate them.
This results in clumsy locomotion behaviours, especially when the terrain has discontinuities.

\subsection{Vision-Based Locomotion}

Using vision unlocks locomotion on complex terrain. 
However, developing such a system is challenging because of the high dimensionality of images.
These challenges pushed prior work to favour modular systems to end-to-end ones, essentially dividing the locomotion task in a vision and control sub-task. 
Vision observations are generally used to build a local elevation map~\cite{tsounis2020deepgait,agrawal2021vision, gaertner2021collision, miki2022learning, fu2021coupling, park2015online}. 
Such maps are used to predict motion primitives (in terms of goal velocities~\cite{fu2021coupling}, foothold placement~\cite{tsounis2020deepgait,agrawal2021vision,gaertner2021collision} or leg motion phase~\cite{miki2022learning}), which are tracked by a separate controller module.
A parallel line of work aims to directly predict navigation commands (either in terms of foothold placement of joint velocities) from on-board depth observations~\cite{margolis2021learning, yu2021visuallocomotion, yang2022learning}.
Overall, the aforementioned methods achieve impressive performance in simulation, but did not yet obtain a similar performance in real-world conditions.
One exception is the recent work of Miki et al.~\cite{miki2022learning}, which obtains impressive results in real-world conditions.
However, such method has very high sensing requirements (up to three lidars or eight stereo modules) and computational costs, making it unfesible on smaller scale robots as the one considered in this work.

\subsection{Self-supervised learning}

Early works in self-supervised learning for locomotion mainly used haptic feedback to identify the terrain~\cite{hoepflinger2010haptic, walas2015terrain, valada2017deep, hoffmann2014effect, wu2016integrated}.
Such information can be used to bias motion over easily traversable environments~\cite{walas2015terrain, Wellhausen2019WhereSI}.
Another line of works estimate traversability information (e.g. probability of falling) from local elevation maps and used it for path planning.
The training data for such estimators is generally collected through experience in simulation~\cite{yang2021real, guzzi2020path, chavez2018learning}.
Similar ideas have also been investigated on other robotic platforms~\cite{kahn2021badgr, nava2019learning, guzzi2020path}.
Overall, the aformentioned works generally operate on an high-level abstraction (traditionally, a trajectory), generally tracked by a separate low-level controller module.
In contrast, our work aims to improve the quality of the low-level controller via self-supervised learning, practically increasing the range of traversable terrain through experience.

\section{Conclusion}
We propose CMS, a technique which uses onboard data collected from multiple sensory inputs to continually improve its performance. We demonstrate this idea for the task of visual walking, using time-shifted proprioception for supervision. Our approach show significant improvements in performance with less than 7 minutes of data per run. One limitation of the current system is that it does not improve the motor system in the real world. An interesting venue for future work is to explore the coupling of techniques that additionally improve the motor policy along with CMS for improving the visual system.

\section{Acknowledgement}
This work was supported by the DARPA Machine Common Sense program and by the ONR MURI award N00014-21-1-2801.
We would like to thank Sasha Sax for the helpful discussions, Noemi Aepli for support with media material, and Haozhi Qi for support with the website creation.

\bibliographystyle{IEEEtran}
\bibliography{references}

\end{document}